\documentclass{article}
\usepackage{arxiv}
\usepackage[utf8]{inputenc} 
\usepackage[T1]{fontenc}    
\usepackage{hyperref}       
\usepackage{url}            
\usepackage{booktabs}       
\usepackage{amsfonts}       
\usepackage{nicefrac}       
\usepackage{microtype}      
\usepackage{lipsum}
\usepackage{graphicx}
\usepackage{multirow}
\usepackage{pdfpages}
\usepackage{hyperref}
\usepackage{enumitem}
\graphicspath{ {./figures/} }
\makeatletter
\newcommand{\printfnsymbol}[1]{
  \textsuperscript{\@fnsymbol{#1}}
}

\title{Multi-Dialect Arabic BERT \\ for Country-Level Dialect Identification}

\author{
    Bashar Talafha\thanks{\hspace{0.075cm} Equal contribution.} \quad \quad
    Mohammad Ali\printfnsymbol{1} \quad\quad
    Muhy Eddin Za'ter \quad\quad
    Haitham Seelawi \\ \bf
    \vspace{0.2cm}    
    Ibraheem Tuffaha \quad\quad
    Mostafa Samir \quad\quad
    Wael Farhan \quad\quad
    Hussein T. Al-Natsheh \\
    Mawdoo3 Ltd, Amman, Jordan \\
    \texttt{\{bashar.talafha,mohammad.ali,muhy.zater,haitham.selawi,} \\
    \texttt{ibraheem.tuffaha,mostafa.samir,wael.farhan,h.natsheh\}} \\
    \texttt{@mawdoo3.com} 
}

\begin{document}
\maketitle

\begin{abstract}
Arabic dialect identification is a complex problem for a number of inherent properties of the language itself.
In this paper, we present the experiments conducted, and the models developed by our competing team, Mawdoo3 AI, along the way to achieving our winning solution to subtask 1 of the Nuanced Arabic Dialect Identification (NADI) shared task. 
The dialect identification subtask provides 21,000 country-level labeled tweets covering all 21 Arab countries.
An unlabeled corpus of 10M tweets from the same domain is also presented by the competition organizers for optional use.
Our winning solution itself came in the form of an ensemble of different training iterations of our pre-trained BERT model, which achieved a micro-averaged F1-score of 26.78\% on the subtask at hand.
We publicly release the pre-trained language model component of our winning solution under the name of Multi-dialect-Arabic-BERT model, for any interested researcher out there.
\end{abstract}

\section{Introduction}
\label{sec:introduction}
The term Arabic language is better thought of as an umbrella term, under which it is possible to list hundreds of varieties of the language, some of which are not even mutually comprehensible. Nonetheless, such varieties can be grouped together with varying levels of granularity, all of which correspond to the various ways the geographical extent of the Arab world can be divided, albeit loosely. Despite such diversity, up until recently, such varieties were strictly confined to the spoken domains, with Modern Standard Arabic (MSA) dominating the written forms of communication all over the Arab world. However, with the advent of social media, an explosion of written content in said varieties have flooded the internet, attracting the attention and interest of the wide Arabic NLP research community in the process. This is evident in the number of held workshops dedicated to the topic in the last few years. 

In this paper we present and discuss the strategies and experiments we conducted to achieve the first place in the Nuanced Arabic Dialect Identification (NADI) Shared Task 1 \cite{mageed-etal-2020-nadi}, which is dedicated to dialect identification at the country level. In section \ref{sec:related_work} we discuss related work. This is followed by section \ref{sec:datasets} in which we discuss the data used to develop our model. Section \ref{sec:models} discusses the most significant models we tested and tried in our experiments. The details and results of said experiments can be found in section \ref{sec:experimental_setup}. The analysis and discussion of the results can be obtained in section \ref{sec:results} followed by our conclusions in section \ref{sec:conclusion}. 

\section{Related Work}
\label{sec:related_work}
The task of Arabic dialect identification is challenging. This can be attributed to a number of reasons, including: a paucity of corpora dedicated to the topic, the lack of a standard orthography between and across the various dialects, and the nature of the language itself (e.g. its morphological richness among other peculiarities). To tackle these challenges, the Arabic NLP community has come up with a number of responses. One response was the development of annotated corpora that focus primarily on dialectical data, such as the Arabic On-line Commentary dataset \cite{zaidan2014arabic}, the MADAR Arabic dialect corpus and lexicon \cite{bouamor2018madar}, the Arap-Tweet corpus \cite{zaghouani2018arap}, in addition to a city-level dataset of Arabic dialects that was curated by \cite{abdul2018you}.

Another popular form of response is the organization of NLP workshops and shared tasks, which are solely dedicated to developing approaches and models that can detect and classify the use of Arabic dialects in written text. One example is the Madar shared task \cite{bouamor2019madar}, which focuses on dialect detection at the level of Arab countries and cities.

The aforementioned efforts by the Arabic NLP community, have resulted in a number of publications that explore the application of a variety of Machine Learning (ML) tools to the problem of dialect identification, with varying emphasis on feature engineering, ensamble methods, and the level of supervision involved \cite{salameh2018fine,elfardy2013sentence,huang2015improved,talafha2019mawdoo3}                         .
The past few years have also witnessed a number of published papers that explore the potential of Deep Learning (DL) models for dialect detection, starting with \cite{elaraby2018deep,ali2018character}, who show the enhanced performance that can be brought about through the use of LSTMs and CNNs, all the way to \cite{zhang2019no}, who highlight the potential of pre-trained language models to achieve state of the art performance on the task of dialect detection.

\section{Dataset}
\label{sec:datasets}
\begin{figure}[h!]
    \centering
    \includegraphics[width=0.90\textwidth]{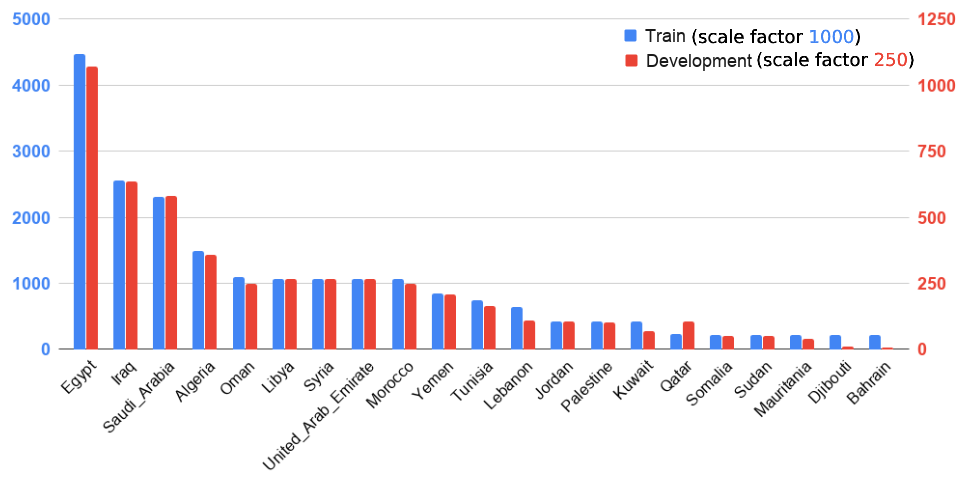}
    \caption{Classes distribution for both Train and Development sets}
    \label{fig:data_distribution}
\end{figure}

The novel dataset of NADI shared task consists of around 31,000 labeled tweets covering the entirety of the 21 Arab countries. Additionally, the task presents an unlabeled corpus of 10M tweets. The labeled dataset is split into 21,000 examples for training, with the rest of the tweets, i.e., 10,000, distributed equally between the development and test sets. Each tweet is annotated with a single country only. In Figure~\ref{fig:data_distribution} we can see the distribution of tweets per country in which \emph{Egypt} and \emph{Bahrain} has the highest and lowest tweet frequencies, respectively. We also note that the ratio of the development to train examples is generally similar across the various dialects, except for the ones with lowest frequencies.

The unlabeled dataset is provided in the form a twitter crawling script, and the IDs of 10M tweets, which in combination can be used to retrieve the text of these tweets. We are able to retrieve 97.7\% of them, as the rest seem to be unavailable (possibly deleted since then or made private). This dataset can be beneficial in multiple ways, including building embedding models (e.g., Word2Vec, FastText)\cite{mikolov2013efficient,bojanowski2017enriching}, pre-training language models, or in semi-supervised learning and data augmentation techniques. This dataset can also be used to further pre-train an already existing language model, which can positively affect its performance on tasks derived from a domain similar to that of the 10M tweets, as we show in our results in Section~\ref{sec:results} 

\section{System Description}
\label{sec:models}
In this section, we present the various approaches employed in our experiments, starting with the winning approach of our Multi-dialect-Arabic-BERT model, followed by the rest of them. The results of our experiments are presented in Table~\ref{tab:results}.
\subsection{Multi-dialect-Arabic-BERT}
\label{sec:bert}
Our top performing approach, which achieved the number one place on the NADI task 1, is based on a Bidirectional Encoder Representations from Transformers (BERT) architecture \cite{devlin2018bert}. BERT uses the encoder part of a Transformer \cite{vaswani2017attention}, and is trained using a masked language model (MLM) objective. This involves training the model to predict corrupted tokens, which is achieved using a special mask token that replaces the original ones. This is typically done on a huge corpus of unlabeled text. The resultant model can then produce contextual vector representations for tokens that capture various linguistics signals, which in turn can be beneficial for downstream tasks.

We started with the \emph{ArabicBERT} \cite{arabic-bert-base}, which is a publicly released BERT model trained on around 93 GB of Arabic content crawled from around the internet. This  model is then fine-tuned on the NADI task 1, by retrieving the output of a special token [CLS],  placed at the beginning of a given tweet. The retrieved vector is in turn fed into a shallow feed-forward neural classifier that consists of a dropout layer, a dense layer, and a softmax activation output function, which produces the final predicated class out of the original 21. It is worth mentioning that during the fine-tuning process, the loss is propagated back across the entire network, including the BERT encoder.

We then were able to significantly improve the results obtained from the model above, by further pre-training \emph{ArabicBERT} on the 10M tweets released by the NADI organizers, for 3 epochs. We refer to this final resultant model as the \emph{Multi-dialect-Arabic-BERT}.

In order to squeeze out more performance from our model, we ended up using ensemble techniques. The best ensemble results came from the voting of 4 models that were trained with different maximum sequence lengths (i.e., 80, 90, 100 and 250). The voting step was accomplished by taking the element-wise average of the predicted probabilities per class for each of these models. The class with the highest value is then outputed as the predicted label.

All of our models were trained using an Adam optimizer \cite{kingma2014adam} with a learning rate of ${3.75}\times{10}^{-5}$ and a batch size of 16 for 3 epochs. No preprocessing was applied to the data except for the processing done by ArabicBERT tokenizer.

We publicly release the Multi-dialect-Arabic-BERT\footnote{\href{https://github.com/mawdoo3/Multi-dialect-Arabic-BERT}{https://github.com/mawdoo3/Multi-dialect-Arabic-BERT}} on GitHub to make it available for use by all researchers for any task including reproducing this paper results.
\subsection{Other Traditional Machine Learning and Deep Learning Models}
In addition to our winning solution, we experimented with a number of other approaches, none of which has exceeded an F1-score of 21, but which we list here for the sake of completeness anyway.
\label{sec:madar-mawdoo3}
\begin{itemize}[noitemsep]
\itemsep 1em 

\item \textit{MADAR-Mawdoo3 Model} \\ Originally proposed by \cite{ragab2019mawdoo3}, three models (i.e. a Multinomial Naive Bayes (MNB), logistic regression, and weak dummy classifier) are trained separately on the data to obtain their dialect probability distributions, which then, in conjunction with TF--IDF vectors, make up the feature space. These features are then fed into an ensamble of five other models (i.e. MNB with one-vs-rest strategy, a Support Vector Machine model (SVM), a Bernoulli Naive Bayes classifier, a K-nearest-neighbours classifier with one-vs rest strategy, and finally a weak dummy classifier). The final predicted classes are obtained using a hard voting approach.

\item \textit{MADAR-Safina Model} \\This model follows Safina model proposed in \cite{bouamor2019madar}. The model is an ensemble of 3 classifiers:Language model classifier based on 5-char n-gram features \cite{heafield-etal-2013-scalable}, Naive Bayes classifier based on 4-to-6 char n-gram features, Naive Bayes classifier based on 1-word n-gram features. The only pre-processing step used is to duplicate every single word for the language model and the char n-gram classifiers. The purpose of duplicating every single word is to detect circumfix n-gram patterns.

\item \textit{MADAR-JUST Model}
\\In this model, we applied the approach proposed by \cite{talafha2019team}. In order to balance the training data, a data augmentation technique based on random shuffling was performed to enlarge and balance the training data. After that, for each sentence, a vector of size 21 that represents a language model probability for each country was extracted and concatenated to a word and character level TF--IDF vectors. An MNB classifier is then applied with the One-vs-the-rest strategy.

\item \textit{FastText Model}
\\FastText \cite{bojanowski2017enriching} was originally implemented to help obtain enhanced word representations over simpler methods such as Word2Vec \cite{mikolov2013efficient}. In our experiments, we pool the FastText vectors of each token in a given sentence, to obtain a fixed-size dense representation of the sentence at hand. This is in turn fed into a multinomial logistic regression for classification \cite{zolotov2017analysis,joulin2016bag}.

\item \textit{Aravec fully connected Model}
\\Aravec is an Arabic based Word2Vec model, trained and published by \cite{soliman2017aravec}. Similar to our FastText model above, we pool the Aravec vectors of the constituent tokens of a sentence to obtain its fixed-size vector representation. However, instead of a conventional ML algorithm, we feed these representations into a feed forward classifier, which is trained to obtain the final predictions \cite{ashi2018pre}.



\end{itemize}

\section{Experiments and Results}
\label{sec:experimental_setup}
\begin{table}[h]
\caption{Final results on NADI development and testing set}
\centering
\begin{tabular}{|l|c|c|c|c|} 
\hline
\multirow{2}{*}{Model}                                 & \multicolumn{2}{c|}{Dev Set Results} & \multicolumn{2}{c|}{Test Set Results}  \\ 
\cline{2-5}
                                                       & Accuracy & F1-Score          & Accuracy & F1-Score            \\ 
\hline
Madar-Safina                                           & 33.35    & 10.1              &    -      &   -                  \\ 
\hline
Logistic-Regression                                    & 35.65    & 16.57             &   -       &   -                  \\ 
\hline
MADAR-1 Mawdoo3                                        & 33.45    & 12.24             &    -      &      -               \\ 
\hline
MADAR-1 JUST                                           & 30.3     & 17.07             &    -      &     -                \\ 
\hline
FastText-embeddings                                    & 34.28    & 19.74             &     -     &     -                \\ 
\hline
Aravec fully connected                                 & 35.67    & 20.86             &     -     &     -                \\ 
\hline
Arabic-BERT-Single                                     & 40.85    & 24.45             &     -     &      -               \\ 
\hline
Arabic-BERT-Ensemble-Diff-Len                          & 41.48    & 24.92             &     -     &      -               \\ 
\hline
Multi-dialect-Arabic-BERT                              & 43.7     & 26                &    -      &    -                 \\ 
\hline
Multi-dialect-Arabic-BERT-Ensemble-Diff-Len            & 44.95    & 27.58             &    \textbf{42.86}      & \textbf{26.78}              \\ 
\hline
Multi-dialect-Arabic-BERT-Ensemble-Diff-Len with rules & \textbf{45.07}    & \textbf{29.03}             &   42.55       & 26.77              \\
\hline
\end{tabular}
\label{tab:results}
\end{table}

As mentioned above, multiple approaches have been investigated in the experiments we conducted, starting with traditional ML techniques then moving to DL approaches, before finally settling on our winning BERT based model. For our traditional ML experiments, we tried various models such as SVM, Logistic Regression (LR) and Naive Bayes (NB), along with features such as TF--IDF. We also tried other ML models that performed well on previous similar tasks such as MADAR Mawdoo3-AI and MADAR Safina models. However, all of these models came short when compared to the BERT models as can be seen in Table \ref{tab:results}, with the best Macro-Averaged F1-score achieved using traditional ML approaches being 17.06\%. 
\begin{table}[h]
\centering
\caption{Final results on NADI testing dataset for the 3 top performing participating teams}
\begin{tabular}{|l|c|c|} 
\hline
\multirow{2}{*}{Model} & \multicolumn{2}{c|}{Results}  \\ 
\cline{2-3} & Accuracy & Macro-Averaged F1-Score \\ 
\hline
ArabicProcessors & 38.34    & 23.26 \\ 
\hline
BERT-NGRAMS & 39.66    & 25.99 \\ 
\hline
Mawdoo3-ai & \textbf{42.86} & \textbf{26.78}  \\ 
\hline
\end{tabular}
\end{table}
We then experimented with a number DL models, along with pre-trained word embedding features, such as FastText and Word2Vec. These models easily surpassed the performance of their traditional ML counterparts, with a maximum macro-averaged F1 score of 20.86\%.

As alluded to above, the best results were achieved by our BERT models. Using the standalone ArabicBERT \cite{arabic-bert-base} we were able to achieve 24.45\% Macro-Averaged F1-score on the development dataset. This score was further increased to 26.46\% using ensemble techniques. This motivated us to further pre-train it on the 10 million unlabelled tweets to form the Multi-dialect-Arabic-BERT model. Using this setup, we were able to achieve a Macro-Averaged Macro-Averaged F1 score of 26\%. Here again, we used the ensemble trick to obtain a 27.58\% Macro-Averaged F1-score on the development set and 26.78\% on the test set, thus winning the competition. We note that applying lexicon-based prediction rules to the best model mentioned above boosted the results of development set to 29.03 F1-score. However, these rules slightly decreased the test set results to 26.77 F1-score, concluding that such rules cause the system to suffer from over-fitting the development set.

\section{Discussion and Analysis}
\label{sec:results}
\begin{figure}[h]
    \centering
    \includegraphics[width=0.50\textwidth]{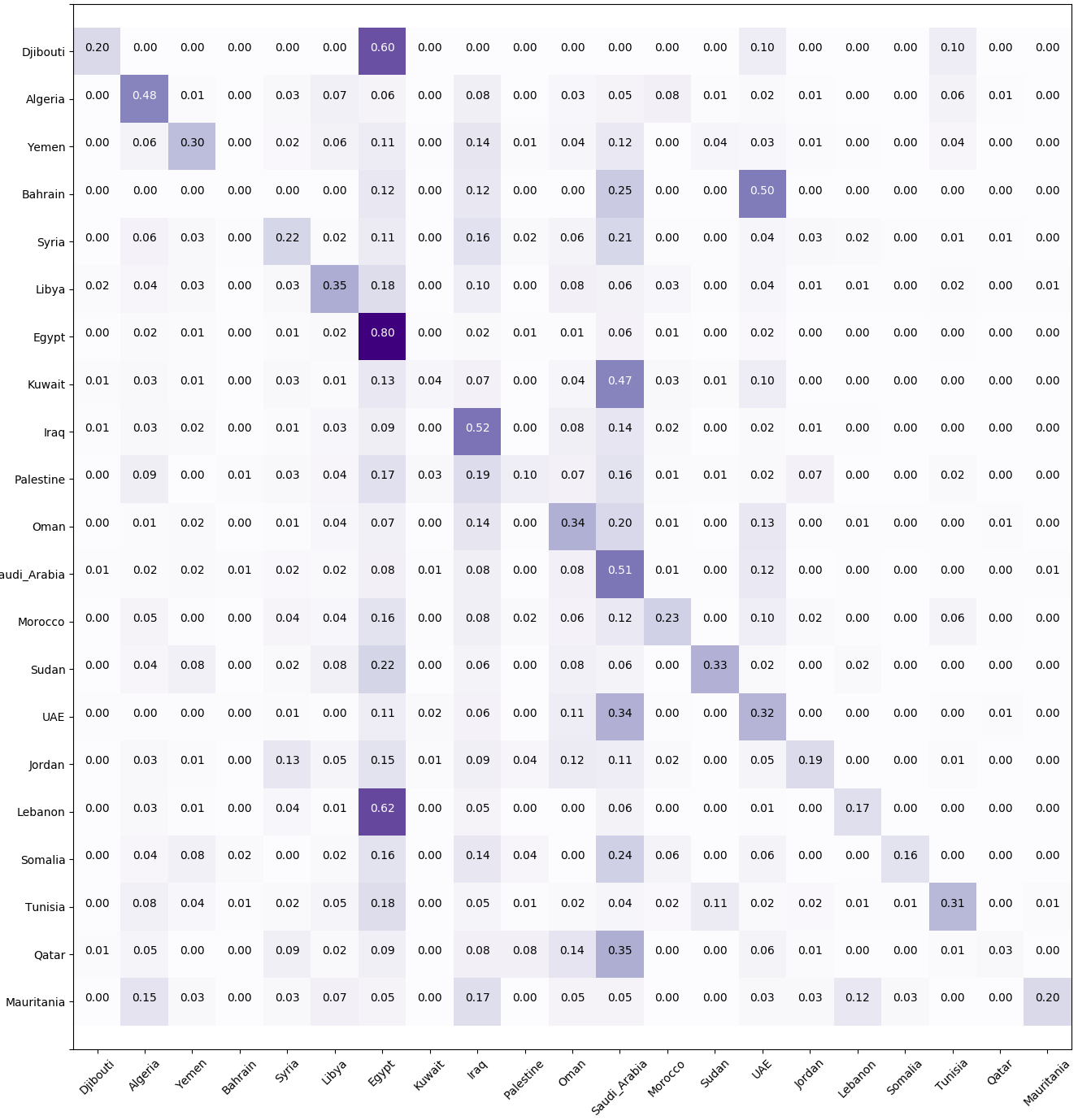}
    \caption{The confusion matrix of our best model on NADI development set}
    \label{fig:cm}
\end{figure}

To aid us with the analysis of the strengths and weaknesses of our winning model, we provide the confusion matrix for its performance on the NADI development set in Figure \ref{fig:cm}.
The matrix highlights a number of issues stemming from the training dataset itself. For instance, it can be clearly seen that the model is biased to the countries with more training data such as Egypt, Iraq and Saudi Arabia; for these countries, the model achieves better results, while achieving much worse F1-scores for the ones with the least training data available.
It can also be seen that the model suffers when trying to differentiate between geographically nearby countries. For example, 50\% of the development samples from Bahrain are labeled as UAE and 22\% from Sudan are labeled as Egypt. This is expected, given the similarities in dialects between neighbouring countries.
Some of the results shown in the confusion matrix have also led us to further investigate the datasets themselves. This resulted in finding that our model does in fact predict the correct class for certain tweets, which were somehow originally mislabeled. Some of these examples can be seen in Table~\ref{tab:incorrect_labels}.

\begin{table}[h]
\caption{Examples of mislabeled and confusing tweets.}
\begin{tabular}{c}
   \includegraphics[page=1,width=.9\textwidth]{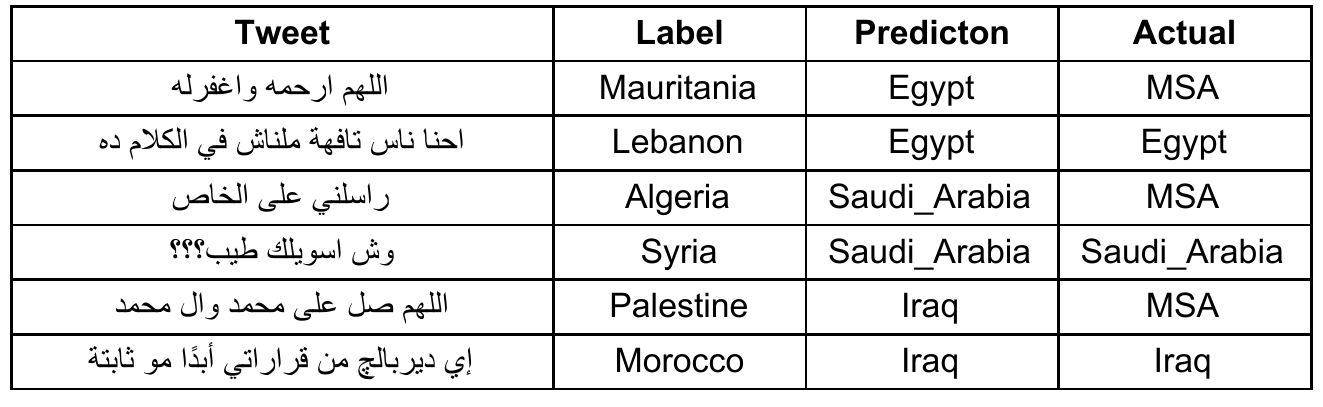}
\end{tabular}
\label{tab:incorrect_labels}
\end{table}

\section{Conclusion}
\label{sec:conclusion}
In this paper we describe our first place solution for the NADI competition, task 1. This was achieved via three stages: firstly, we further pre-trained a publicly released BERT model (i.e. Arabic-BERT) on the 10 millions tweets supplied by the NADI competition organizers. Secondly, we trained the resultant model on the NADI labelled data for task 1, multiple times, independently, with each of these iterations using a different mixture of maximum sentence length and learning rate. Thirdly, we selected the 4 best performing iterations (based on their performance on the development dataset), and aggregated their softmax predictions via a simple element wise averaging function, to produce the final prediction for a given tweet. For future work, we would like to investigate other advanced pre-training methods, such as XLNET, and ELECTRA, which we believe might hold the key to better performance on this task.

\bibliographystyle{unsrt}  
\bibliography{references}

\end{document}